\documentclass{ecai}
\usepackage{times}
\usepackage{graphicx}
\usepackage{latexsym}
\usepackage{amsmath}
\usepackage{amssymb}
\begin{document}

\title{Encoding Cryptographic Functions to SAT Using \textsc{Transalg} System\footnote{Short variant of this paper was accepted to the ECAI2016 conference.}}

\author{Ilya Otpuschennikov$^2$, Alexander Semenov$^2$, Irina Gribanova$^2$, Oleg Zaikin$^2$, Stepan Kochemazov \institute{Matrosov Institute for System Dynamics and Control Theory of SB RAS,
Irkutsk, Russia, email: otilya@yandex.ru, biclop.rambler@yandex.ru, the42dimension@gmail.com, zaikin.icc@gmail.com, veinamond@gmail.com} }

\maketitle
\bibliographystyle{ecai}

\begin{abstract}
In this paper we propose the technology for constructing propositional encodings of discrete functions. It is aimed at solving inversion problems of considered functions using state-of-the-art SAT solvers. We implemented this technology in the form of the software system called \textsc{Transalg}, and used it to construct SAT encodings for a number of cryptanalysis problems. By applying SAT solvers to these encodings we managed to invert several cryptographic functions. In particular, we used the SAT encodings produced by \textsc{Transalg} to construct the family of two-block MD5 collisions in which the first 10 bytes are zeros. Also we used \textsc{Transalg} encoding for the widely known A5/1 keystream generator to solve several dozen of its cryptanalysis instances in a distributed computing environment. In the paper we compare in detail the functionality of \textsc{Transalg} with that of similar software systems.
\end{abstract}

\section{INTRODUCTION}
In the recent years, there appeared many works on cryptanalysis, which use various automated software systems. These include software systems aimed at solving cryptanalysis equations using generic algorithms as well as systems that generate such equations based on descriptions of considered cryptographic functions.

Among the promising methods for solving equations corresponding to cryptanalysis problems we would like to note algebraic cryptanalysis \cite{Bard:2009:AC:1618541} and also actively developing SAT-based cryptanalysis \cite{Cook97findinghard}. The latter is based on the state-of-the-art algorithms for solving Boolean satisfiability problem (SAT). These algorithms are successfully applied to combinatorial problems from various areas \cite{DBLP:series/faia/2009-185}. Also the cryptanalysis of some cryptographic functions used in the real world (such as A5/1 keystream generator and hash functions from the MD family) was performed via SAT. 

One of the main problems that arises when one wants to apply SAT to cryptanalysis of a particular cipher, consists in obtaining a SAT encoding for a corresponding algorithm. A lot of cryptographic primitives have a number of specific properties that make this process quite nontrivial. Therefore it is relevant to develop software systems for automating construction of SAT encodings of cryptographic functions. In the recent years several systems of that sort appeared.

In \cite{Otpuschennikov2015} we described the \textsc{Transalg} software system developed by us specifically for solving inversion problems of cryptographic functions via SAT. Note that in \cite{Otpuschennikov2015} due to limitations on the volume of the paper we could not include any results of computational experiments on the use of \textsc{Transalg} for solving cryptanalysis problems. In the present paper we describe new results on inversion of several cryptographic functions that were obtained with the use of \textsc{Transalg} system and compare \textsc{Transalg} functionality with that of several similar systems.

Let us give a brief outline of the paper. In the second section we introduce main notions regarding SAT and describe the ideas underlying SAT-based cryptanalysis. In the third section we briefly describe the \textsc{Transalg} system. In the fourth section we compare encodings produced by \textsc{Transalg} with that constructed by other similar systems. Also in this section we briefly compare the effectiveness of SAT and SMT solvers in application to cryptanalysis instances and conclude that SAT solvers nowadays are more effective in this particular area. In the fifth section we use \textsc{Transalg} encodings to perform SAT-based cryptanalysis of ciphering systems used in the real world. In particular, we describe the cryptanalysis of the A5/1 generator in the distributed computing environment and show the results on the search for collisions of MD4 and MD5 hash functions.

\section{THEORETICAL FOUNDATIONS OF SAT-BASED CRYPTANALYSIS}

Boolean Satisfiability Problem (SAT) consists in the following: for an arbitrary Boolean formula to decide whether it is satisfiable or not. It can be effectively (in polynomial time on the size of an original formula) reduced to SAT for a formula in the Conjunctive Normal Form (CNF). Usually in the context of SAT it is assumed that we consider formula in this form. Since SAT is NP-complete problem, today there are no algorithms that could solve SAT in polynomial time. However, there is a number of heuristic algorithms that show good effectiveness in application to various practical problems.  In the recent years the scope of application of these algorithms has greatly increased \cite{DBLP:series/faia/2009-185}.

By $\{0,1\}^{*}$ we denote the set of all binary words of an arbitrary finite length. By discrete functions we mean arbitrary (possibly, partial) functions of the kind:
\begin{equation}
\label{eq1_f}
f:\{0,1\}^{*}\rightarrow \{0,1\}^{*}.
\end{equation}
Hereinafter we consider only total computable discrete functions. In other words we assume that $f$ is specified by some program (algorithm) $A_f$, that has finite runtime on each word from $\{0,1\}^{*}$. The program $A_f$ specifies a family of functions of the kind $f_n:\{0,1\}^n\rightarrow\{0,1\}^{*}$, $n\in N_1$. The problem of inversion of an arbitrary function $f_n$ is formulated as follows: based on the known $y\in Range\;f_n$ and the known algorithm $A_f$, find such $x\in\{0,1\}^n$ that $f_n(x)=y$.

Many cryptanalysis problems can be formulated as inversion problems of discrete functions. For example, suppose that given a secret key $x\in\{0,1\}^n$, $f_n$ generates  a pseudorandom sequence (generally speaking, of an arbitrary length), that is later used to cipher some plaintext via bit-wise XOR. Such a sequence is called a keystream. Knowing some fragment of  plaintext lets us know the corresponding fragment of  keystream, i.e. some word $y$ for which we can consider the problem of finding such $x\in\{0,1\}^n$, that $f_n(x)=y$. Regarding cryptographic keystream generators this corresponds to the so called \textit{known plaintext attack}.

Let us give another example. Total functions of the kind $f:\{0,1\}^{*}\rightarrow\{0,1\}^c$, where $c$ is some constant are called \textit{hash functions}. If $n$ is the length of the input message and $n>c$, then there exist such $x_1,x_2$, $x_1\neq x_2$, that $f_n(x_1)=f_n(x_2)$. Such a pair $x_1,x_2$ is called a collision of a hash function $f$. A cryptographic hash function is considered compromised if one is able to find collisions of that function in reasonable time.

The functions of the kind \eqref{eq1_f} that are used in cryptography are usually constructed in such a way that their inversion problems are computationally hard. Meanwhile the functions themselves should be computed fast, otherwise the ciphering speed of the cryptosystem based on such a function would be low. The ability to effectively compute $f$ makes it possible to apply to its inversion the approach based on the propositional encoding of $A_f$ program. As a result we reduce the inversion problem for an arbitrarty function of the kind $f_n$ to the problem of finding a satisfying assignment of some satisfiable CNF, as it follows from the Cook-Levin theorem \cite{DBLP:conf/stoc/Cook71}. Below we describe the basic idea how such CNFs are constructed using Boolean circuits representing the functions of the kind $f_n$.

Let us consider some function $f_n:\{0,1\}^n\rightarrow\{0,1\}^m$ defined by a program $A_f$. Analyzing $A_f$ we can construct Boolean circuit $S(f_n)$ over some complete basis $B$, such that $S(f_n)$ implements $f_n$. Hereinafter we assume that $B=\{\neg,\wedge\}$. The $S(f_n)$ circuit has $n$ inputs and $m$ outputs. It can be considered as marked directed acyclic graph. In this graph $n$ selected nodes are marked with Boolean variables $x_1,\ldots,x_n$. These nodes correspond to the circuit inputs and form the set $X$. All the other nodes are marked with elements from basis $B$. Among them $m$ nodes correspond to circuit outputs.

For $S(f_n)$ circuit in polynomial time on its size we can construct CNF $C(f_n)$. For this purpose all the circuit nodes that are not input we mark with variables, that form the set $V$, $X\bigcap V=\emptyset$. The variables from $V$ are called auxiliary variables. With each logical gate $G$ of the circuit $S(f_n)$ we therefore assign some auxiliary variable $v(G)\in V$. There is the subset $Y$, $Y\subseteq V$, $Y=\{y_1,\ldots,y_m\}$ formed by the variables corresponding to $S(f_n)$ outputs. 

Let $v(G)$ be an arbitrary variable from $V$ and $G$ be a corresponding gate. If $G$ is a NOT-gate and $u\in X\bigcup V$ is the variable linked with the input of $G$ then we encode the gate $G$ with the Boolean formula $v(G)\leftrightarrow \neg u$ (hereinafter by $\leftrightarrow$ we mean logical equivalence). If $G$ is an AND-gate and $u,w\in X\bigcup V$ are variables linked with the gate inputs then we encode $G$ with the formula $v(G)\leftrightarrow u\wedge w$. CNF-representations of Boolean functions specified by formulas $v(G)\leftrightarrow \neg u$ and $v(G)\leftrightarrow u\wedge v$ look as follows (respectively):
\begin{equation}
\label{eq2_f}
\begin{array}{l}
\left(v(G)\vee u\right)\wedge \left(\neg v(G)\vee \neg u\right)\\
\left(v(G)\vee \neg u\vee \neg w\right)\wedge \left(\neg v(G)\vee u \right)\wedge \left(\neg v(G)\vee w\right)
\end{array}
\end{equation}
Thus with an arbitrary gate $G$ of circuit $S(f_n)$ we associate CNF $C(G)$ of the kind \eqref{eq2_f}. We will say that CNF $C(G)$ encodes gate $G$, and CNF 
\begin{equation}
\label{eq3_f}
C(f_n)=\bigwedge\limits_{G\in S(f_n)}C(G)
\end{equation}
encodes circuit $S(f_n)$. The described technique of constructing a CNF for a circuit $S(f_n)$ is known as Tseitin transformations \cite{Tseitin83}. To a CNF $C(f_n)$ of the kind \eqref{eq3_f} we will refer as \textit{template CNF} representing the algorithm that implements $f_n$.

Let $x$ be an arbitrary Boolean variable. Below we will use the following notation: by $l_\beta(x)$, $\beta\in\{0,1\}$ we denote the literal $\neg x$ if $\beta=0$, and literal $x$ if $\beta=1$. Let $(\beta_1,\ldots,\beta_m)$ be an arbitrary assignment from $Range f_n\subseteq \{0,1\}^m$. From the properties of the Tseitin transformations it follows that CNF 
\begin{equation}
\label{eq4_f}
C(f_n)\wedge l_{\beta_1}(y_1)\wedge\ldots\wedge l_{\beta_m}(y_m)
\end{equation}
is satisfiable and from each assignment satisfying \eqref{eq4_f} we can effectively extract such an assignment $(\alpha_1,\ldots,\alpha_n)\in \{0,1\}^n$ that $f_n(\alpha_1,\ldots,\alpha_n)=(\beta_1,\ldots,\beta_m)$. Thus we reduced the inversion problem for $f_n$ to SAT for a CNF of the kind \eqref{eq4_f}.

From the above SAT-based cryptanalysis looks as follows: first we reduce an inversion problem for a considered cryptographic function to SAT and then we apply SAT solving technologies to a SAT instance obtained. In the next section we describe the software system called \textsc{Transalg}, that was designed and developed by us specifically to produce SAT encodings for cryptographic functions.

\section{BRIEF DESCRIPTION OF \textsc{TRANSALG} SOFTWARE SYSTEM}

Ideologically the \textsc{Transalg} system is built upon the principles of symbolic execution \cite{King:1976:SEP:360248.360252}. The discrete functions are specified by the programs for the abstract machine. \textsc{Transalg} applies standard techniques from the compilation theory to these programs. However, as a result of translation it produces not an executable code but a Boolean formula that encodes a considered algorithm (in the form of template CNF).

To describe discrete functions the \textsc{Transalg} system uses the domain specific language called the TA-language. The TA-language has a C-like syntax and block structure. An arbitrary block (composite operator) is essentially a list of instructions, and has its own (local) scope. In the TA-language one can use nested blocks with no limit on depth. During the analysis of a program \textsc{Transalg} constructs a scope tree with the global scope at its root. Every identifier in a TA-program belongs to some scope. Variables and arrays declared outside of any block and also all functions belong to the global scope and therefore can be accessed in any point of a program.

A TA-program is a list of functions. The \texttt{main} function is the entry point and, thus, must exist in every program. The TA-language supports basic constructions used in procedural languages (variable declarations, assignment operators, conditional operators, loops, function calls, etc.), various integer operations and bit operations including bit shifting and comparison.

The main data type in the TA-language is the \texttt{bit} type. \textsc{Transalg} uses this type to establish links between variables used in a TA-program and Boolean variables included into a corresponding propositional encoding. It is important to distinguish between these two sets of variables. Below we will refer to variables that appear in a TA-program as \textit{program variables}. All variables included in a propositional encoding are called \textit{encoding variables}. 

Given a TA-program $A(f_n)$ as an input, \textsc{Transalg} constructs the propositional encoding of the function $f_n$. Below we will refer to this process as the translation of the TA-program $A(f_n)$. Essentially, the translation of the TA-program $A(f_n)$ is the symbolic execution of this program \cite{King:1976:SEP:360248.360252}.

Upon the translation of an arbitrary instruction that contains a program variable of the \texttt{bit} type, \textsc{Transalg} links this program variable with the corresponding encoding variable. \textsc{Transalg} establishes such links only for program variables of the \texttt{bit} type. Variables of other types, in particular \texttt{int} and \texttt{void} are used only as service variables, e.g. as loop counters or to specify functions that do not return value.

Declarations of global \texttt{bit} variables can have an \texttt{\_\_in} or an \texttt{\_\_out} attribute. The \texttt{\_\_in} attribute marks variables that contain input data for an algorithm. The \texttt{\_\_out} attribute marks variables that contain an output of an algorithm. Local \texttt{bit} variables cannot be declared with these attributes.

The translation of a TA-program has two main stages. At the first stage, \textsc{Transalg} parses the source code of this TA-program and constructs a syntax tree using standard techniques of the compilation theory. At the second stage, the system performs symbolic execution of a TA-program to construct the corresponding propositional encoding.

The process of symbolic execution of a TA program can be divided into elementary steps. As a result of each elementary step a new encoding variable $x$ is introduced and the following formula is constructed
\begin{equation}
\label{eq5_f}
x\leftrightarrow g\left(\tilde{x}_1,\ldots,\tilde{x}_k\right),
\end{equation}
in which $\tilde{x}_1,\ldots,\tilde{x}_k $ are some encoding variables introduced earlier (by “$\leftrightarrow$” we denote the logical equivalence).
The propositional encoding of a TA program is a set of formulas of the kind \eqref{eq5_f}. 

Cryptographic algorithms often use various bit shifting operators and also copy bits from one cell to another without changing their value. During the symbolic execution of such operators there may appear elementary steps, producing the formulas of the kind $x\leftrightarrow \tilde{x}$. However, we do not really need such formulas in the propositional encoding since it is evident that without the loss of correctness we can replace an arbitrary formula of the kind $x'\leftrightarrow g(x,\ldots)$ by formula $x'\leftrightarrow g(\tilde{x},\ldots)$. 
In other words it is not necessary to introduce the encoding variable $x$. \textsc{Transalg} tracks such situations using special data structures. The corresponding technique is described in detail in \cite{Otpuschennikov2015}. 

The \textsc{Transalg} system has full support of conditional operators. Let $\Phi(z_1,\ldots,z_k)$ be an arbitrary expression of the TA language, where $z_1,\ldots, z_k$ are program variables of the \texttt{bit} type. Assume that these variables are linked with encoding variables $x_1,\ldots,x_k$, and that Boolean formula $\phi(x_1,\ldots,x_k)$ is the result of translation of an expression $\Phi(z_1,\ldots,z_k)$. Below we say that expression $\Phi(z_1,\ldots,z_k)$ is associated with Boolean formula $\phi(x_1,\ldots,x_k)$.

Let us consider the BNF form of the conditional operator. 
\begin{verbatim}
<if_statement> := if (<expr>) <statement> 
		[else <statement>]
\end{verbatim}
Here \texttt{<expr>} is a predicate of a conditional operator. Assume \texttt{<expr>} is $\Phi$ and expression $\Phi$ is associated with a formula $\phi$. Let $\delta_1$ and $\delta_2$ be Boolean formulas associated (in the aforementioned sense) with the expressions $\Delta_1$ and $\Delta_2$ of the TA-program.
%Let $\Delta_1$ and $\Delta_2$ be expressions of the TA-program, that are associated with formulas $\delta_1$ and $\delta_2$, respectively. 
Without the loss of generality assume that in then- and else-branches we have assignment operators $z=\Delta_1$ and $z=\Delta_2$, where $z$ is some program variable linked with an encoding variable $x$. Then during the translation of this conditional operator \textsc{Transalg} creates a new encoding variable $x'$, links it with $z$ and adds the following formula to the propositional encoding:
\begin{equation}
\label{eq6_f}
x'\leftrightarrow \phi \wedge \delta_1 \vee \neg \phi \wedge \delta_2.
\end{equation}
If there is no assignment $z=\Delta_2$ in the else-branch, or if there is no else-branch, then \eqref{eq6_f} transforms into $x'\leftrightarrow\phi\wedge\delta_1\vee\neg\phi\wedge x$. Likewise, if there is no assignment $z=\Delta_1$ in the then-branch, then \eqref{eq6_f} transforms into $x'\leftrightarrow\phi\wedge x\vee\neg\phi\wedge\delta_2$.

During the translation of TA-program for function $f_n$ to SAT the \textsc{Transalg} system represents this function as a composition of functions with smaller number of inputs. If $f_n$ is represented as a composition of only basis functions (for example, from the basis $\{\wedge,\neg\}$) then to construct the corresponding CNF one can use only Tseitin transformations. 
%However, \textsc{Transalg} can represent $f_n$ as a composition of more complex (compared to basis) functions. 
%It makes it possible to reduce the number of variables in the final SAT encoding. In these cases 
However, \textsc{Transalg} can represent $f_n$ as a composition of functions over more than two Boolean variables. To encode corresponding functions to SAT \textsc{Transalg} employs the truth tables. Also in these cases \textsc{Transalg} uses Boolean minimization. On the current stage it employs the \textsc{Espresso}\footnote{http://embedded.eecs.berkeley.edu/pubs/downloads/Espresso/index.htm} Boolean minimization library. This choice is motivated by the fact that this library for decades remains one of de facto standards in the area. The \textsc{Espresso} was embedded into \textsc{Transalg} as one of its modules.

\section{COMPARISON OF \textsc{TRANSALG} WITH SIMILAR SYSTEMS}

Note, that, generally speaking, it is possible to construct SAT encodings for discrete functions using various symbolic verification systems, such as CBMC \cite{ckl2004}. However, such systems are not designed specifically for the purpose of solving cryptanalysis problems. It means that in practice when they are used to produce such propositional encodings, there arise a number of issues, e.g. big runtime, inability to correctly interpret results since sets of input and output variables are unknown, etc. Because of this it is considered relevant to develop software systems specifically for constructing SAT encodings for cryptanalysis problems. In the paper we consider three systems that have more or less similar functionality to that of \textsc{Transalg}. Below we briefly describe these systems. 

The \textsc{Grain of Salt} system is designed to produce SAT encodings only for cryptographic keystream generators based on the shift registers. In \cite{DBLP:conf/tools/Soos10} this system was used to construct propositional encodings for the Bivium, Trivium and Grain keystream generators. Unfortunately, \textsc{Grain of Salt} does not work with conditional operators and therefore can not be used to encode a number of cryptographic functions (for example the A5/1 generator).

The URSA system \cite{journals/lmcs/predrag} is a generic propositional encoding tool that is applicable to a wide class of combinatorial problems, varying from CSP (Constraint Satisfaction Problem) to cryptography. To describe these problems \textsc{URSA} uses proprietary domain specific language. To solve SAT instances produced, \textsc{URSA} uses two embedded solvers: \textsc{argosat} and \textsc{clasp}.

The \textsc{Cryptol} system \cite{DBLP:conf/csiirw/ErkokM09,DBLP:conf/plpv/ErkokM09} is designed as a tool for analysis of cryptographic specifications using SMT solvers. It uses functional Haskell-like domain specific language to describe the algorithms. Also the authors of \textsc{Cryptol} made the \textsc{SAW} (Software Analysis Workbench\footnote{http://saw.galois.com/}) tool that makes it possible to produce SAT and SMT encodings for cryptographic problems defined in \textsc{Cryptol} language.

\subsection{Comparison of SAT and SMT approaches to inversion of cryptographic fucntions}

The question whether SAT or SMT solvers are better suited for solving cryptanalysis instances is quite controversial. That is why we compared SAT and SMT in application to cryptanalysis of two keystream generators. The first is the strengthened version of the Geffe generator \cite{Geffe} (essentially that is a special case of the Threshold generator \cite{Bruer}). We considered the variant of this generator that uses three Linear Feedback Shift registers (LFSRs) defined by the following primitive polynomials:
$$
\begin{array}{c}
x^{31}+x^7+1\\
x^{32}+x^7+x^5+x^3+x^2+x+1\\
x^{33}+x^{16}+x^4+x+1\\
\end{array}
$$
Thus the secret key for this generator has the length of 96 bits. Its registers are shifted synchronously and the keystream bit is defined as follows:
\begin{equation}
\label{geffe}
(x_1\wedge x_2)\oplus (x_2\wedge x_3)\oplus (x_1\wedge x_3)
\end{equation}
where by $x_i$, $i\in \{1,2,3\}$ we mean the most significant bit of the i-th LFSR (instead of \eqref{geffe} the standard Geffe generator uses the function $ (x_1\wedge x_2)\oplus (x_2\wedge x_3)\oplus x_3$). We considered the cryptanalysis problem in the following form: to find 96-bit secret key using the known keystream fragment of length 200 bits. 

In a similar way we studied the cryptanalysis of the Bivium keystream generator \cite{DBLP:conf/isw/Canniere06}. In particular, we considered this problem in the following formulation: to find initial values of 177 cells of generator registers, based on analysis of 200 bits of keystream. To make test instances solvable in reasonable time we assumed that some bits of secret key are already known (i.e. used the so-called guessing bits in terms of \cite{Bard:2009:AC:1618541}). In particular we constructed tests with 30 guessing bits. Below we refer to this cryptanalysis problem as \textit{Bivium30}.

At the current moment \textsc{Transalg} can not output encoding in SMT format, therefore to construct SMT encodings for considered functions we used only \textsc{Cryptol+SAW}. The corresponding SAT instances were generated using both \textsc{Cryptol+SAW} and \textsc{Transalg}. For each problem outlined above we constructed 100 SMT instances generated by \textsc{Cryptol+SAW}, 100 SAT instances generated by \textsc{Cryptol+SAW} and 100 SAT instances constructed by \textsc{Transalg}. To solve SMT instances we used high-ranked SMT solvers \textsc{Boolector}, \textsc{Yices}, \textsc{CVC4} and \textsc{Z3}. The \textsc{Boolector} solver showed the best results among SMT solvers. To solve SAT instances we used widely known \textsc{Minisat} solver \cite{DBLP:conf/sat/EenS03}. Specifically we used \textsc{Minisat 2.2}. The results of the experiments are shown in Table \ref{sat-vs-smt_table}. The time limit for Geffe was 1 minute, time limit for Bivium was 1 hour. The solvers were launched on one core of AMD Opteron 6276.

\begin{table}
\begin{center}
{\caption{Comparison between SAT and SMT}\label{sat-vs-smt_table}}
		\begin{tabular}{|l|c|c|c|}
		\hline				
		& \multicolumn{2}{|c|}{\textsc{Cryptol+SAW}}&\textsc{Transalg}\\
		\cline{2-3}
		& SMT&SAT&\\
		\hline
		\multicolumn{4}{|c|}{Strengthened Geffe} \\
		\hline
		Solver & \textsc{Boolector} & \textsc{Minisat} & \textsc{Minisat} \\
		\hline
		Solved & 81 & 100 & 100 \\
		\hline
		Avg. time, sec. & 35.69 & 7.69 & 7.14 \\
		\hline
		\hline
		\multicolumn{4}{|c|}{Bivium30}\\
		\hline
		Solver & \textsc{Boolector} & \textsc{Minisat} & \textsc{Minisat} \\
		\hline
		Solved & 54 & 83 & 81 \\
		\hline
		Avg. time, sec. & 1 744 & 1 165 & 1 262 \\
		\hline
		\end{tabular}
		\end{center}
\end{table}

%From the presented results it can be seen that SMT solvers are significantly less effective than SAT solvers for cryptanalysis even if we consider relatively simple cryptographic functions. Therefore below we concentrate on SAT encodings for cryptanalysis problems. 
From the presented results we can conclude that state-of-the-art SMT solvers lose to SAT solvers if we consider their applications to solving inversion problems of cryptographic functions. However, we would like to point out that in this paper we do not consider the questions regarding the construction of special Background theories, that would take into account the features of considered cryptanalysis problems. Moreover, we currently are not aware if there exist such Background theories. It is possible that in future the developments in this direction will make it possible to drastically improve the performance of SMT solvers on cryptanalysis instances.

\subsection{Comparison of SAT encodings}
In the next series of experiments we considered the Bivium, Trivium and Grain keystream generators. For these generators we constructed SAT encodings using the \textsc{Grain of Salt}, URSA, \textsc{Cryptol+SAW} and \textsc{Transalg} systems. 

Note that SAT-based cryptanalysis of Bivium, Trivium and Grain was studied for example in \cite{DBLP:conf/SAT/EibachPV08,DBLP:conf/tools/Soos10,DBLP:conf/SAT/SoosNC09}. In accordance with these papers we considered the inversion problems for the following functions:
\begin{equation*}
f^{Bivium}:\{0,1\}^{177}\rightarrow\{0,1\}^{200}, f^{Grain}:\{0,1\}^{160}\rightarrow\{0,1\}^{160}.
\end{equation*}
Cryptanalysis of Trivium was considered as the inversion problem for function 
\begin{equation*}
f^{Trivium}:\{0,1\}^{288}\rightarrow\{0,1\}^{300}.
\end{equation*}
The parameters of the obtained encodings are shown in Table \ref{encodings_table}.

\begin{table}
\begin{center}
{\caption{The parameters of encodings for Bivium, Trivium and Grain}\label{encodings_table}}	
\begin{tabular}{|l|r|r|r|r|}
			\hline
			& \textsc{GoS} & \textsc{URSA} & \textsc{Cryptol+SAW} & \textsc{Transalg} \\
			\hline
			\multicolumn{5}{|c|}{\textsc{Bivium}} \\
			\hline
			Vars & 842 & 1 637 & 1 432 & 442 \\
			\hline
			Clauses & 6 635 & 5 975 & 5 308 & 7 960 \\
			\hline
			Literals & 29 455 & 16 995 & 15 060 & 40 320 \\
			\hline
			\hline
			\multicolumn{5}{|c|}{\textsc{Trivium}} \\
			\hline
			Vars & 1 887 & 4 284 & 3 097 & 1 587 \\
			\hline
			Clauses & 22 881 & 15 885 & 11 889 & 22 176 \\
			\hline
			Literals & 118 413 & 45 657 & 34 037 & 109 872 \\
			\hline
			\hline
			\multicolumn{5}{|c|}{\textsc{Grain}} \\
			\hline
			Vars & 4 546 & 9 279 & 4 246 & 1 785 \\
			\hline
			Clauses & 74 269 & 37 317 & 16 522 & 34 165 \\
			\hline
			Literals & 461 069 & 105 925 & 46 402 & 190 388 \\
			\hline
		\end{tabular}
		\end{center}
\end{table}

In the next series of experiments we used the constructed SAT encodings for Bivium, Trivium and Grain to solve a number of cryptanalysis instances for the corresponding keystream generators, assuming that a number of bits from the secret key are known. These bits we will below refer to as \textit{guessing bits} \cite{Bard:2009:AC:1618541}. In other words assume that we consider  the inversion problem of function $f_n:\{0,1\}^n\rightarrow\{0,1\}^m$ in some point $y\in Range\;f_n$. Let $C(f_n)$ be the template CNF for $f_n$ and let $X^{in}$ be the set of variables encoding the input of $f_n$. Then for the inversion problem for $f_n$ in point $y$ we assume that for some subset $X'\subseteq X^{in}$ the values of variables from $X'$ in the preimage $x\in\{0,1\}^n$:$f_n(x)=y$ are known. A SAT encoding for this modified inversion problem for $f_n$ can be produced from $C(f_n)$ by assigning known values to all variables from $X'$.

For the cryptanalysis problems for Bivium, Trivium and Grain in the formulation described above we considered SAT encodings constructed using all the aforementioned software systems. Then for each generator we modified these encodings by fixing the values of variables from the set $X'\subseteq X^{in}$. By \textit{GeneratorK} we mean the SAT instances which encode cryptanalysis of the corresponding generator, modified by assigning values to variables from some set $X'$, $|X'|=K$.

Essentially, \textit{GeneratorK} means a series of SAT instances that differ in values of variables from $X'$. We considered such series of 100 instances each. On instances from each series we ran the CDCL SAT solvers that rated high in SAT competition 2014 \cite{sc2014proceedings}: \textsc{ROKK}, \textsc{MiniSat\_blbd}, \textsc{MiniSat\_HACK\_999ED}, \textsc{SWdia5by} and \textsc{Riss Blackbox}. In case of the encodings produced by \textsc{URSA} we were forced to use only the solvers \textsc{clasp} and \textsc{ArgoSAT} embedded into this system. To solve each instance each solver had a time limit of one hour. During the analysis of the obtained experimental data for each series of the kind \textit{GeneratorK} we chose the best solver judging by the amount of SAT instances solved within the time limit. The corresponding information is shown in Table \ref{solving_table}. Also in this table we show the average time in seconds, that the best solver demonstrated on solved tests.

\begin{table}
\begin{center}
{\caption{Solving cryptanalysis instances for Bivium, Trivium and Grain with guessing bits (avg. time computed for solved within time limit)}\label{solving_table}}	
		\begin{tabular}{|l|r|r|r|r|}	
			\hline
			& \textsc{GoS} & \textsc{URSA} & \textsc{Cryptol+SAW} & \textsc{Transalg} \\
			\hline
			\multicolumn{5}{|c|}{\textsc{Bivium30}} \\
			\hline
			Solver & \textsc{ROKK} & \textsc{clasp} & \textsc{ROKK} & \textsc{ROKK} \\
			\hline
			Solved & 100 & 97 & 100 & 99 \\
			\hline
			Avg. time, sec. & 1 037 & 1 415 & 1 188 & 1 042 \\
			\hline
			\hline
			\multicolumn{5}{|c|}{\textsc{Trivium142}} \\
			\hline
			Solver & \textsc{ROKK} & \textsc{clasp} & \textsc{ROKK} & \textsc{ROKK} \\
			\hline
			Solved & 97 & 100 & 99 & 94 \\
			\hline
			Avg. time, sec. & 1 691 & 1 406 & 1 462 & 1 429 \\
			\hline
			\hline
			\multicolumn{5}{|c|}{\textsc{Grain102}} \\
			\hline
			Solver & \textsc{ROKK} & \textsc{clasp} & \textsc{ROKK} & \textsc{ROKK} \\
			\hline
			Solved & 59 & 41 & 56 & 77 \\
			\hline
			Avg. time, sec. & 2 007 & 1 689 & 1 479 & 1 589 \\
			\hline
		\end{tabular}
		\end{center}		
\end{table}

As a final note we would like to discuss how \textsc{Transalg} differs from other systems. 
The distinctive feature of \textsc{Transalg} is that it can construct and explicitly output the template CNF $C(f_n)$. When it constructs $C(f_n)$ it employs the concept of symbolic execution of program $A_f$ fully reflecting this process in the memory of abstract computing machine. As a result, in a template CNF $C(f_n)$ all elementary operations with the memory of abstract machine are represented in the form of Boolean equations over sets of Boolean variables. \textsc{Transalg} makes it possible to work with these variables directly, thus providing a number of useful features for cryptanalysis. In particular, we can quickly generate families of cryptographic instances: to make certain SAT instance for function inversion it is sufficient to add to a template CNF unit clauses encoding the corresponding output. That is why template CNFs are very handy when one uses partitioning strategy \cite{Hyvarinen11} to solve some hard SAT instance in a distributed computing environment. Also \textsc{Transalg} can identify variables corresponding to inputs and outputs of considered function, so external tools can be used to check correctness of SAT encodings and to analyze the results of solving SAT. In particular, thanks to this we can use any SAT solvers and preprocessors. \textsc{Transalg} allows to monitor the values of program variables inside program $A_f$ at any step of computing, and, therefore, to assert any conditions on these variables. For example, thanks to this it is easy to write in a program $A_f$ the conditions specifying the differential path for finding collisions of cryptographic hash functions. In other considered systems (URSA, Cryptol) there arise significant difficulties when writing such conditions. Finally, let us note that the connection between the structure of CNF $C(f_n)$ and an original algorithm, reflected by \textsc{Transalg}, can play an important role in implementation of several cryptographic attacks (such as guess-and-determine attacks \cite{Bard:2009:AC:1618541}) in parallel.

\section{INVERSION OF REAL WORLD CRYPTOGRAPHIC FUNCTIONS USING \textsc{TRANSALG}}

In this section we present our results on SAT-based cryptanalysis of several ciphering systems, that continue to be used in practice despite being compromised. First we describe the SAT-based cryptanalysis of the A5/1 keystream generator. Then we present our results on finding collisions of MD4 and MD5 cryptographic hash functions. In all these cases we used the encodings produced by the \textsc{Transalg} system.

\subsection{SAT-based cryptanalysis of A5/1}

A5/1 is probably one of the most widely known keystream generators. Despite the fact that in various sources there were described several attacks on it (with different degrees of success), it is still used in many countries to cipher the GSM data.
The description of the A5/1 algorithm can be found, for example, in \cite{DBLP:conf/fse/BiryukovSW00}. 

We considered the cryptanalysis problem for the A5/1 generator in the following formulation. 
We assume that we know first 114 bits of keystream\footnote{In the GSM protocol the messages are transmitted in blocks of 114 bits called bursts.} that were generated by the generator from 64-bit secret key. We need to find the secret key.
Therefore, essentially we need to solve the inversion problem for the function
\begin{equation}
\label{a51func}
f^{A5/1}:\{0,1\}^{64}\rightarrow\{0,1\}^{114}
\end{equation}
specified by the A5/1 algorithm. Let $C(f^{A5/1})$ be the template CNF for this function and let $X^{in}=\{x_1,\ldots,x_{64}\}$ be the set of variables encoding the input of $f^{A5/1}$. By assigning values from an arbitrary $y\in Range\;f^{A5/1}$ to the corresponding output variables in $C(f^{A5/1})$ we produce the CNF $C_y(f^{A5/1})$.

The SAT instances encoding the inversion of function \eqref{a51func} are very difficult for sequential SAT solvers. In \cite{DBLP:conf/pact/SemenovZBP11} there was described the method that can be used to solve such SAT instances in parallel. According to that method to solve the original cryptanalysis instance one has to solve $2^{31}$ simplified SAT instances in the worst case scenario. Each of these simplified instances is produced as a result of assigning values to variables from the special set $\tilde{X}$, $\tilde{X}\subset X^{in}$, $|\tilde{X}|=31$ in the CNF $C_y(f^{A5/1})$. The set $\tilde{X}$ is called the decomposition set and its structure is shown in \cite{DBLP:conf/pact/SemenovZBP11}. All the simplified SAT instances produced this way form a partitioning for $C_y(f^{A5/1})$. The aforementioned features of \textsc{Transalg} make it possible to effectively outline an arbitrary decomposition set in a SAT encoding. After this each instance from a SAT partitioning, produced by set $\tilde{X}$, is constructed by adding to CNF $C_y(f^{A5/1})$ the corresponding unit clauses. These instances can be solved in parallel independently from each other. Using this SAT partitioning strategy we solved several dozens of cryptanalysis instances for A5/1 in a special distributed computing system. In detail the corresponding experiment is described in \cite{Semenov2016}.

Additionally we would like to note that thanks to the knowledge of the set $X^{in}$ in the SAT encoding and the use of external SAT solver we managed to significantly improve the effectiveness of cryptanalysis. In particular, we modified the \textsc{Minisat 2.2} solver (in which we increased starting activity for the variables from $X^{in}$ and changed values of parameters \texttt{var\_decay} and \texttt{clause\_decay}). Without this we would have to  spend up to 100 times more computing resources on solving corresponding cryptanalysis problems.

%We would like to point out that \textsc{Transalg} allows us to know which Boolean variables in constructed CNFs correspond to inputs and outputs of considered functions. This fact made it possible to introduce modifications to the \textsc{Minisat 2.2} solver (in which we increased starting activity for the variables from $X^{in}$ and changed values of parameters \texttt{var\_decay} and \texttt{clause\_decay}). Without this we would have to  spend up to 100 times more computing resources on solving corresponding cryptanalysis problems.

\subsection{SAT-based approach to finding collisions of hash functions from MD family}

To construct SAT encodings for finding collisions of hash functions MD4 and MD5 we used only the \textsc{Transalg} system because of the reasons outlined in the end of Section 4. 

Let $f:\{0,1\}^*\rightarrow\{0,1\}^c$ be some cryptographic hash function, that works with messages split into blocks of length $n$, $n>c$. It defines the function of the kind $f_n:\{0,1\}^n\rightarrow\{0,1\}^c$. To produce the SAT encoding for the problem of finding collisions of this function we essentially translate the program describing $f_n$ twice, using disjoint sets of Boolean variables. Let $C_1$ and $C_2$ be the corresponding CNFs in which the sets of output variables are $Y^1=\left\{y_1^1,\ldots, y_c^1\right\}$ and $Y^2=\left\{y_1^2,\ldots, y_c^2\right\}$. Then finding collisions of $f_n$ is reduced to finding an assignment that satisfies the following Boolean formula:
\begin{equation}
\label{eq4}
C_1\wedge C_2\wedge \left( y_1^1 \leftrightarrow y_1^2 \right) \wedge \ldots \wedge \left( y_c^1 \leftrightarrow y_c^2 \right).
\end{equation}

Note, that \eqref{eq4} can be extremely hard even for state-of-the-art SAT solvers. This is particularly the case for the hash functions of the MD family \cite{DBLP:conf/crypto/Rivest90}. To make this problem solvable in realistic time we added to SAT for \eqref{eq4} the constraints, that arise in differential attacks on the considered hash functions \cite{DBLP:conf/eurocrypt/WangLFCY05,DBLP:conf/eurocrypt/WangY05}. Note that the first implementation of a similar attack on MD4 and MD5 in the SAT form was described in \cite{DBLP:conf/sat/MironovZ06}. To produce the corresponding SAT encodings the authors of \cite{DBLP:conf/sat/MironovZ06} used the encoding technique developed specifically for the considered hash functions. In Table \ref{MD_table} we compare characteristics of the SAT instances for finding collisions of MD4 and MD5, that were used in \cite{DBLP:conf/sat/MironovZ06}, with that of encodings produced with the \textsc{Transalg} system.

\begin{table}
\begin{center}
{\caption{The parameters of encodings for hash functions MD4 and MD5}\label{MD_table}}
		\begin{tabular}{|c|c|c|c|}	
			\hline
					& & Encodings from \cite{DBLP:conf/sat/MironovZ06} & \textsc{Transalg}\\
			\hline
			MD4 & variables & 53 228  & 19 363 \\
			\cline{2-4}
					& clauses   & 221 440 & 184 689\\
			\hline
			MD5 & variables & 89 748  & 35 477 \\
			\cline{2-4}
					& clauses   & 375 176 & 304 728\\
			\hline				
		\end{tabular}
		\end{center}
\end{table}

First, we considered the problem of finding single block collisions of the MD4 hash function (taking into account the differential paths from \cite{DBLP:conf/eurocrypt/WangLFCY05}). The authors of \cite{DBLP:conf/sat/MironovZ06} note that in their implementation of attack on MD4 it took about 500 seconds to find one collision. Using the SAT encodings produced by \textsc{Transalg} and \textsc{Cryptominisat} solver \cite{DBLP:conf/SAT/SoosNC09} we managed to find about 1000 MD4 collisions within 500 seconds on one core of Intel i5-3570T (4 Gb RAM).

After this we studied the problem of finding two-block collisions of MD5. Note that the hash functions of the MD family are based on the Merkle-Damgard construction \cite{DBLP:conf/crypto/Merkle89,DBLP:conf/crypto/Damgard89a} and work with the messages split into 512-bit blocks. More precisely, the hash value is computed iteratively for the message split into $N$ 512-bit blocks $M_1,\ldots,M_N$. Let us denote by $h_j$ the hash value computed at the iteration number $j\in\{1,\ldots,N\}$. Then according to the Merkle-Damgard construction, $h_j=f^{MD}(h_{j-1},M_j)$. The value $h_0$ is called the Initial Value (IV) and is usually fixed in the algorithm specification. The process of constructing the two-block collision of MD5 has two stages. In the first stage we search for two 512-bit blocks $M_1$ and $M'_1$, for which the difference between hash values modulo $2^{32}$ is equal to \texttt{(0x80000000, 0x82000000, 0x82000000, 0x82000000)}, in accordance with the constraints on the differential path from \cite{DBLP:conf/eurocrypt/WangY05}. These constraints as well as a number of additional constraints referred to in \cite{DBLP:conf/eurocrypt/WangY05} as \textit{bit conditions} were added to the resulting SAT encoding. We denote $h_1=f^{MD5}(IV,M_1)$, $h'_1=f^{MD5}(IV,M'_1)$. In the second stage we look for second 512 blocks $M_2$ and $M'_2$ such that $f^{MD5}(h_1,M_2)=f^{MD5}(h'_1,M'_2)$. 

The SAT instances encoding the search for $M_1$ and $M'_1$ turned out to be too difficult for the majority of the solvers that we tested. We were able to obtain consistent results for these problems only using \textsc{Plingeling} and \textsc{Treengeling} solvers \cite{DBLP:conf/sat/Biere14} (versions from the SAT competition 2014 \cite{sc2014proceedings}). We used a computing cluster, each computing node of which contains 2 AMD Opteron 6276 processors (32 processor cores in one node). On each computing node we ran one copy of \textsc{Plingeling} or \textsc{Treengeling} in multi-threaded mode (32 threads). On different computing nodes we ran several such copies in parallel. The solving time on different nodes varied significantly: from several hours to several days.

During experiments we noticed the following interesting phenomenon: the \textsc{Treengeling} solver found several blocks $M_1$ and $M'_1$ with a lot of zeroes in the beginning. A more detailed analysis showed that if we assign first 10 bytes in $M_1$ and $M'_1$ with $0$s then the corresponding SAT instance is satisfiable, but if in addition to this we assign $0$ to the 11-th byte, then the CNF becomes unsatisfiable (and the solver gives the corresponding answer quite fast). Thus we outlined the class of pairs of the kind $M_1$ and $M'_1$ that satisfy the differential path from \cite{DBLP:conf/eurocrypt/WangY05}, and have first 10 zero bytes. In 89 hours using 15 cluster nodes (480 cores) we managed to find 20 pairs of the described kind. For the obtained pairs $(M_1,M'_1)$  the problem of constructing such pairs $(M_2,M'_2)$ that the messages $M_1|M_2$ and $M'_1|M'_2$ form the two-block collision of MD5, turned out to be relatively simple: on average one such pair $(M_2,M'_2)$ was found by the \textsc{Plingeling} solver in 400 seconds on one cluster node. An example of the collision of this kind is shown in Table \ref{MD5_collision_example}.

\begin{table}
\begin{center}
{\caption{MD5 two-block collisions with 10 zero bytes in the beginning}\label{MD5_collision_example}}
%\scriptsize{
\begin{tabular}{|l|l|}
\hline
%--1
&		00 00 00 00 00 00 00 00 00 00 e0 5c 2f 3c f5 48 \\
&32 1e cc \underline{a0} bf 25 b9 bd ed 93 8d 88 c3 c9 f5 e4\\
&		55 2d 34 05 06 c6 b3 00 9b f4 b2 83 75 \underline{71} fa 1e \\
&f3 26 84 73 04 57 ab 23 0e ca 73 \underline{02} d6 5b a3 aa\\
$M$&	54 4f 48 19 c2 3d b1 f4 12 2b 6e 8d 9f 31 40 ad \\
&c6 f4 66 \underline{99} fc 02 44 dd 14 09 a0 47 d0 c8 5d af\\
&		c1 bf b6 6e 51 d7 f5 87 d6 81 32 d8 93 \underline{00} \underline{e4} dd \\
&0f 59 e5 6b 96 f9 9b e4 13 df 64 \underline{ae} 90 69 b6 a6\\
\hline
&		00 00 00 00 00 00 00 00 00 00 e0 5c 2f 3c f5 48\\
&32 1e cc \underline{20} bf 25 b9 bd ed 93 8d 88 c3 c9 f5 e4\\
&		55 2d 34 05 06 c6 b3 00 9b f4 b2 83 75 \underline{f1} fa 1e\\
&f3 26 84 73 04 57 ab 23 0e ca 73 \underline{82} d6 5b a3 aa\\
$M'$&54 4f 48 19 c2 3d b1 f4 12 2b 6e 8d 9f 31 40 ad \\
&c6 f4 66 \underline{19} fc 02 44 dd 14 09 a0 47 d0 c8 5d af\\
&		c1 bf b6 6e 51 d7 f5 87 d6 81 32 d8 93 \underline{80} \underline{e3} dd \\
&0f 59 e5 6b 96 f9 9b e4 13 df 64 \underline{2e} 90 69 b6 a6\\
\hline
$H$&178477e15fde4ff267aa55438d539b16\\
\hline
\end{tabular}
%}
\end{center}
\end{table}

In conclusion we would like to once more point out the features of \textsc{Transalg} system that made it possible to obtain the presented results. It is mainly thanks to the translation concept of \textsc{Transalg} that allows one to directly work with variables encoding each elementary step of considered algorithm. That is why we can effectively reflect in SAT encoding any additional constraints, such as, for example, the ones that specify a differential path. In similar software systems this step requires a significant amount of work to be implemented.

\section{RELATED WORK}

The research on applying combinatorial algorithms to solving cryptanalysis problems has been actively conducted for the recent 10-15 years. The idea to use cryptanalysis instances as hard tests for Boolean satisfiability solvers was first expressed in \cite{Cook97findinghard}. The work \cite{DBLP:journals/jar/MassacciM00} contains one of the first examples of propositional encodings of cryptographic functions. It should be noted that the SAT instances from \cite{DBLP:journals/jar/MassacciM00} and several other later papers turned out to be too difficult for SAT solvers and therefore did not allow  researchers to perform the cryptanalysis of the corresponding cryptographic systems. The work \cite{DBLP:conf/sat/MironovZ06} was the first successful attempt to apply SAT to cryptanalysis instances for the relevant (at that moment) ciphering systems.

The monograph \cite{Bard:2009:AC:1618541} studies various aspects of algebraic cryptanalysis. Quite significant part of this work contains the results on application of SAT solvers to solving algebraic equations over finite fields.

%The monograph \cite{Bard:2009:AC:1618541} contains systematic research of various questions regarding algebraic cryptanalysis. A substantial part of this book studies the possibilities of the use of SAT solvers for solving cryptanalysis equations represented in the form of algebraic systems over finite fields.

In our opinion, cryptanalysis of keystream generators using the SAT approach is quite promising area of research, because increasing the speed of stream ciphering often leads to the decrease of complexity of inversion problem of the corresponding function. A number of papers studied the application of SAT to cryptanalysis of keystream generators. In \cite{Mcdonald_attackingbivium} the Bivium cipher was studied using the \textsc{Minisat} solver. Later this direction of research was developed in \cite{DBLP:conf/SAT/EibachPV08}. In \cite{DBLP:conf/SAT/SoosNC09} the abilities of the \textsc{Cryptominisat} solver in application to cryptanalysis of several keystream generators (including Bivium) were demonstrated. In \cite{DBLP:conf/pact/SemenovZBP11} using SAT in a distributed environment several cryptanalysis instances of the A5/1 keystream generator were solved. 

In \cite{DBLP:conf/pact/SemenovZ15} there was described the method for automated search for SAT partitionings and its application to finding new partitionings for the SAT instances encoding the cryptanalysis of A5/1 ang Bivium generators. Using these partitionings it was possible to improve the estimations for Bivium that were presented in \cite{DBLP:conf/SAT/SoosNC09}. 

The first example of application of SAT to finding collisions of cryptographic hash functions was proposed in \cite{DBLP:conf/frocos/JovanovicJ05}. In \cite{DBLP:conf/sat/MironovZ06} for the first time the collisions of cryptographic hash functions from the MD family were constructed using SAT. The key idea of that paper, which made it possible to succeed, was to augment the SAT encodings of the considered hash functions with additional Boolean constraints encoding the differential paths from \cite{DBLP:conf/eurocrypt/WangLFCY05,DBLP:conf/eurocrypt/WangY05}. In \cite{DBLP:conf/SAT/DeKV07,jsatFJD} there are some results on application of SAT to inversion attacks on cryptographic hash functions from the MD family.

In recent years the SAT community has developed many encoding techniques that can be applied to a wide class of combinatorial problems. A lot of references to key papers in this area can be found in \cite{DBLP:series/faia/Prestwich09}. There is a number of systems for automated encoding of Constraint Satisfaction Problem to SAT (for example, \cite{DBLP:conf/cpaior/HebrardOO10,DBLP:conf/sat/SohTB13}, among others). However, the automated systems that can effectively encode cryptographic functions to SAT (if we take into account various aspects that are specific for cryptography) are rare. Apart from \textsc{Transalg} these are the \textsc{Grain of Salt} \cite{DBLP:conf/tools/Soos10}, \textsc{URSA} \cite{journals/lmcs/predrag} and \textsc{Cryptol+SAW} \cite{DBLP:conf/csiirw/ErkokM09,DBLP:conf/plpv/ErkokM09} tools. The detailed comparison of the mentioned systems with \textsc{Transalg} can be found in Section 4 of the present paper.

\section{CONCLUSIONS AND FUTURE WORK}

In the present paper we introduced the \textsc{Transalg} system designed to automatically translate the inversion problems of discrete functions to SAT. We used \textsc{Transalg} to solve inversion problems of several cryptographic functions. The source code of the \textsc{Transalg} system can be found in the repository\footnote{https://gitlab.com/transalg/transalg}. Examples of TA-programs and the corresponding template CNFs constructed using \textsc{Transalg}, and also all cryptanalysis tests we studied in Sections 4-5 and found MD5 collisions of described kind are available in the repository\footnote{https://gitlab.com/satencodings/satencodings}.

In our opinion, with the use of \textsc{Transalg} we clearly demonstrated the practical applicability of SAT-based cryptanalysis. We believe that this direction of research is very promising and intend to obtain new interesting results on this path. In particular, we plan to apply SAT-based cryptanalysis to inversion of several hash functions. From our point of view it is possible to achieve the results similar to that of \cite{Dobbertin:1998:FTR:647933.740752}: that some hash functions with truncated number of rounds are not one-way. Also we are going to study the effectiveness of SAT-based cryptanalysis in application to finding collisions of hash functions using various differential paths, which are different from that found in \cite{DBLP:conf/eurocrypt/WangLFCY05,DBLP:conf/eurocrypt/WangY05}. For example, it will be interesting to construct and study the SAT encodings based on the differential paths from \cite{DBLP:journals/iacr/Stevens12}.

\ack
Authors express deep gratitude to Predrag Jani\v{c}i\'{c} and Aaron Tomb for feedback regarding several important details of the \textsc{URSA} system and \textsc{Cryptol} system, respectively. We also thank Karsten Nohl for his help with the Rainbow method implementation in the A5/1 Cracking Project and Alexey Ignatiev for valuable advices and helpful discussions that made it possible to significantly improve the quality of the paper. 

The research was funded by Russian Science Foundation (project no 16-11-10046).

\bibliography{refs}

\end{document}